\documentclass[
]{ceurart}

\usepackage{listings}
\usepackage{booktabs}
\begin{document}

\copyrightyear{2025}
\copyrightclause{Copyright © 2025 for this paper by its authors. Use permitted under Creative Commons License Attribution 4.0 International (CC BY 4.0).}

\conference{MiGA@IJCAI25: International IJCAI Workshop on 3rd Human Behavior Analysis for Emotion Understanding, August 29, 2025, Guangzhou, China.}

\title{CLIP-MG: Guiding Semantic Attention with Skeletal Pose Features and RGB Data for Micro-Gesture Recognition on the iMiGUE Dataset}



\author[1]{Santosh Patapati}[%
orcid=0009-0007-4780-2624,
email=santosh@cyrionlabs.org,
]

\author[1]{Trisanth Srinivasan}[%
orcid=0009-0009-7588-7498,
email=trisanth@cyrionlabs.org,
]

\author[1]{Amith Adiraju}[%
email=aadiraju@cyrionlabs.org,
]

\address[1]{Cyrion Labs, Texas, United States}




\begin{abstract}
  Micro-gesture recognition is a challenging task in affective computing due to the subtle, involuntary nature of the gestures and their low movement amplitude. In this paper, we introduce a Pose-Guided Semantics-Aware CLIP-based architecture, or CLIP for Micro-Gesture recognition (CLIP-MG), a modified CLIP model tailored for micro-gesture classification on the iMiGUE dataset. CLIP-MG integrates human pose (skeleton) information into the CLIP-based recognition pipeline through pose-guided semantic query generation and a gated multi-modal fusion mechanism. The proposed model achieves a Top-1 accuracy of 61.82\%. These results demonstrate both the potential of our approach and the remaining difficulty in fully adapting vision-language models like CLIP for micro-gesture recognition.
\end{abstract}


\maketitle

\section{Introduction}

Micro-gestures (MGs) are subtle, spontaneous body movements that can reveal hidden emotional states \cite{cohn2000micro,funes2019review}, often occurring when people attempt to suppress their true feelings. Unlike overt actions or expressive gestures, micro-gestures involve minute motions (e.g., slight fidgeting, brief facial or limb movements) that are short in duration and low in amplitude, making them hard to detect and classify \cite{pantic2009affective}. The analysis of micro-gestures has gained traction in affective computing and human behavior understanding because these involuntary cues provide valuable insight into a person's internal state. Automatic recognition of micro-gestures is therefore important for applications in psychology, human-computer interaction, and emotion analysis \cite{kapoor2007automatic, patapati2024trillm}.

In this paper, we present CLIP-MG, a novel multi-modal framework for micro-gesture classification on iMiGUE. Our approach builds upon previous work by incorporating pose (skeleton) data in a principled way. The main contributions are summarized as follows:
\begin{enumerate}
    \item We develop a system that uses human pose cues to help guide the semantic query extraction from video frames. The skeleton information helps focus the CLIP visual encoder on the regions of subtle motion. This creates a semantic query embedding that is rich with features relevant to pose.
    \item We introduce a gated fusion mechanism to combine visual and skeleton representations effectively. Our gated fusion learns to weight and integrate the two modalities. This allows pose features to adaptively modulate the visual features before and during the cross-attention process.
    \item We extend CLIP to a multi-modal setting. The pose-based query is fed into the CLIP transformer for cross-attention over semantically significant visual token features. This limits the model to attend to parts relevant to gesture. This results in a fused representation that has both semantic and motion-specific information for classification.
    \item We evaluate CLIP-MG on the iMiGUE micro-gesture dataset. We additionally perform numerous ablation studies to quantify how much each proposed component improves performance. The results of our ablation studies provide insights for future researchers and future research directions.
\end{enumerate}

\section{Related Works}
\subsection{The iMiGUE Dataset}
Recent progress in this area has been driven by the introduction of specialized datasets for micro-gesture understanding. In particular, iMiGUE is a large-scale video dataset introduced by Liu et al. \cite{liu2021imigue} for identity-free micro-gesture understanding and emotion analysis. The iMiGUE dataset consists of video footage of tennis players during post-match interviews, with detailed frame-level annotations of various micro-gestures. The dataset contains 72 subjects (split into 37 for training and 35 for testing in a cross-subject protocol) and a total of 33 gesture classes.

One thing to note is that the class distribution is highly imbalanced. 28 of the 33 classes are tail classes with relatively few samples, meaning they collectively only make up less than 60\% of the data. This long-tailed distribution, combined with the subtlety and high intra-class variability of micro-gestures, makes the recognition task extremely challenging \cite{miga2024overview}.

\subsection{Signal Processing \& Machine Learning Techniques for Micro-Gesture Recognition}
To tackle these challenges, the research community has organized the Micro-Gesture Analysis (MiGA) challenges in recent years \cite{miga2024overview, miga2023overview}. These competitions have spurred the development of novel multi-modal approaches that leverage both video (RGB) and skeleton (pose) modalities for micro-gesture recognition. In the 2024 MiGA Challenge, for example, all top-performing methods integrated pose information alongside RGB frames. The winning entry by Chen et al. introduced a prototype-based learning approach with a two-stream 3D CNN (PoseConv3D) backbone for RGB and pose, cross-modal attention fusion, and a prototypical refinement component to calibrate ambiguous samples \cite{chen2024prototype}. This method achieved a Top-1 accuracy of 70.25\% on the iMiGUE test set, substantially outperforming earlier approaches. The second-place method by Huang et al. proposed a multi-scale heterogeneous ensemble network (M2HEN) combining a 3D convolutional model and a Transformer for feature diversity, reaching 70.19\% accuracy \cite{huang2024m2hen}. Another notable approach by Wang et al. leveraged the vision-language model CLIP: they used a frozen CLIP as a teacher network for RGB frames and injected CLIP-derived text embeddings into a pose-based model, achieving 68.9\% accuracy with an ensemble of RGB, joint, and limb pose streams. These efforts demonstrate that multi-modal fusion and semantic knowledge transfer are highly important in improving micro-gesture recognition \cite{wang2024clipdistill}.

\subsection{CLIP-Based Video Understanding}
Meanwhile, in the broader action recognition field, researchers have explored using pre-trained vision-language models like CLIP for video understanding. CLIP (Contrastive Language-Image Pre-training) \cite{radford2021clip} has shown powerful visual feature representations aligned with semantics via natural language supervision. However, straightforward fine-tuning of CLIP on video data can neglect smaller semantic information. To address this, Quan et al. proposed Semantic-Constrained CLIP (SC-CLIP) \cite{quan2025scclip}. SC-CLIP adapts CLIP to video by generating a compact semantic query from dense visual tokens and using cross-attention to refocus the model on those action-relevant semantics. This "constrains" CLIP’s attention to discriminative features and yields stronger zero-shot and fine-grained recognition.

SC-CLIP demonstrates that directing attention to semantically meaningful regions can improve fine-grained video understanding. Micro-gestures, despite being low in their extent of movement, still take place with subtle visual semantics. Skeleton key-points give precise spatial-temporal anchors (hands, face, shoulders) that show where and when these cues take place. Thus, we design a pose-guided semantic attention mechanism that uses skeletal cues to steer CLIP towards where the gesture is taking place. This creates a query that captures the subtle semantics of micro-gestures.

\section{Methodology}
\begin{figure*}[!t]
    \centering
    \includegraphics[width=1\linewidth]{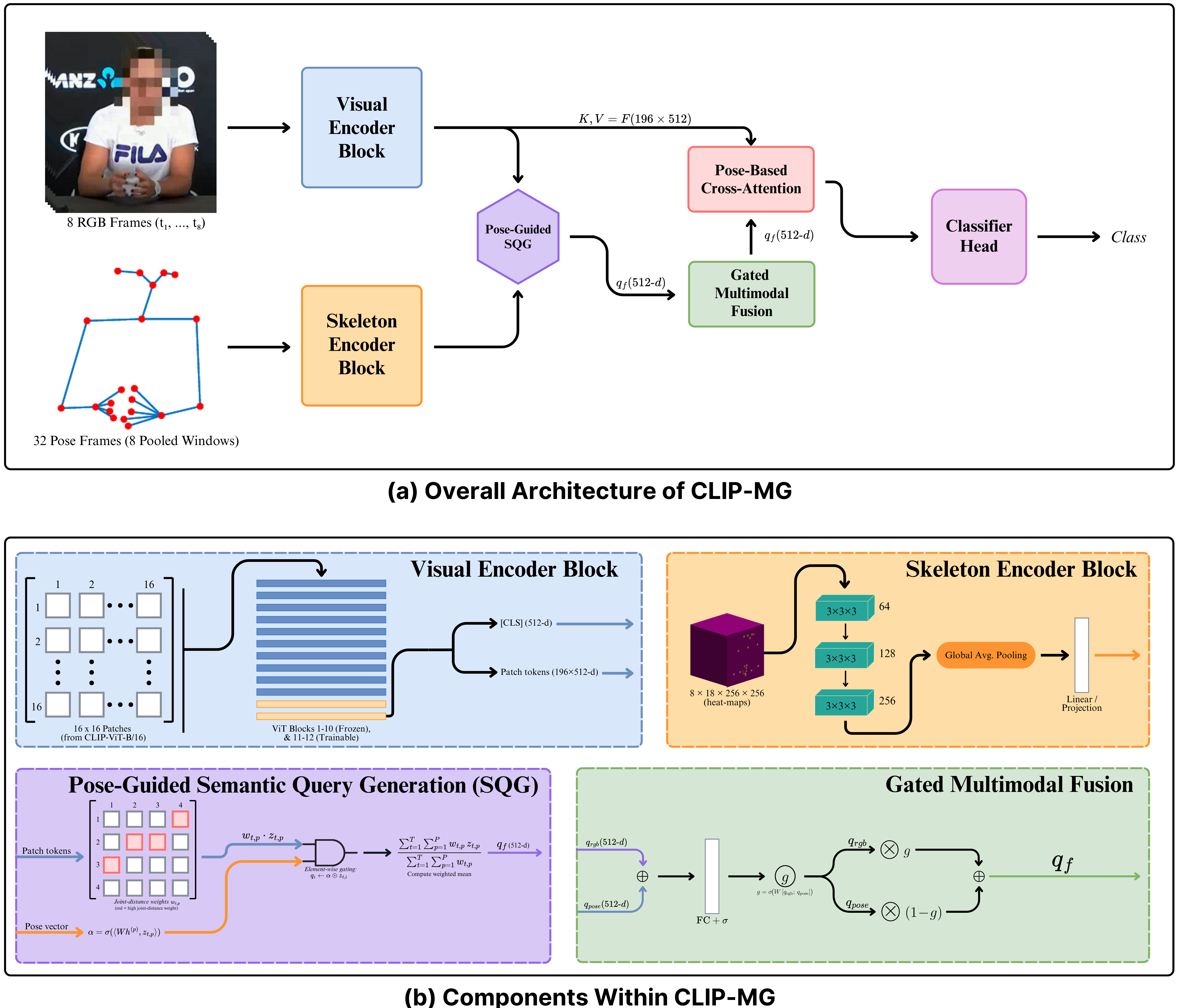}
    \noindent
    \caption{
    \noindent
    Overview of the proposed architecture for micro-gesture classification. The CLIP-MG pipeline integrates a frozen CLIP ViT-B/16 visual encoder and an OpenPose-based skeleton encoder to generate pose-guided semantic queries. This focuses attention on image patches relevant to gestures. A gated multimodal fusion and cross-attention mechanism then blend the visual and pose features before a simple classification head predicts the micro-gesture label.
    }
    \label{fig:Architecture}
\end{figure*}

Our model (illustrated in Figure 1) has several components working in sequence: (1) a visual encoder (based on CLIP’s vision transformer) processes the RGB frames, (2) a skeleton encoder processes the pose sequence, (3) pose-based semantic query generation producse semantic queries from visual features guided by pose features, (4) a gated fusion and semantics-based cross-attention fuses the modalities and improves the representation, and (5) a classification head outputs the predicted gesture label. In the following, we detail each component and the overall pipeline.

\subsection{Visual Encoder}
We adopt the OpenAI CLIP ViT-B/16 image tower \cite{dosovitskiy2020vit} with the standard $224 × 224$ input resolution and $16 × 16$ patching, which yields $P = 196$ patch tokens plus one [CLS] token per frame. The internal transformer width is 768 dimensions, while CLIP's projection head maps the final [CLS] embeddings to a 512-dimensional space. From each micro-gesture clip we uniformly sample $T'=8$ frames. Formally, for frame $t$ we obtain the token sequence:

\[
Z_t = \{\,z_{t,\mathrm{CLS}},\;z_{t,1},\;\dots,\;z_{t,196}\,\},
\qquad
z_{t,\mathrm{CLS}}\in\mathbb{R}^{768}\,.
\]

During training we freeze the first 10 of the 12 ViT blocks and fine-tune only the last two blocks together with our added components \cite{frozenclip2023}. Temporal information is aggregated by average-pooling the eight [CLS] embeddings to produce the per-clip visual feature.

\subsection{Skeleton Encoder}
We use the OpenPose format \cite{cao2017openpose} to extract skeleton features. Given a clip, we first sample $T' = 32$ pose frames:

\[
X^{(p)} = \bigl\{x^{(p)}_{1},\,\dots,\,x^{(p)}_{32}\bigr\},
\qquad
x^{(p)}_{i}\in\mathbb{R}^{18\times 2}\,.
\]

To stay time-aligned with the eight RGB frames, the 32 pose frames are grouped into eight non-overlapping four-frame windows centered on \(t_1,\dots,t_8\). The heat-maps of each window are average-pooled along the temporal axis. This results in eight pose volumes that correspond one-to-one with the RGB inputs.

Each joint is then rasterized into a $256 \times 256$ canvas as a 2D Gaussian \cite{zhou2019hpeheatmap}:
\[
H^{(t)}_{n,i,j}
\;=\;
\exp\!\Biggl(
  -\frac{(i - x^{(t)}_{n})^{2} + (j - y^{(t)}_{n})^{2}}{2\,\sigma^{2}}
\Biggr),
\qquad
\sigma = 2.5\text{\,px},
\]
where $(x^{(t)}_{n},\,y^{(t)}_{n})$ is the $n$-th joint of frame $t$.  
Stacking all joints and time-steps yields a 4D tensor
\[
\mathcal{H}\;\in\;\mathbb{R}^{8 \times 18 \times 256 \times 256}\,.
\]

Utilizing an implementation similar to Tessa \cite{patapati2025tessa}, we then employ a three-stage 3x3x3 convolutional network (with channel depths of 64, 128, and 256) to encode subtle pose dynamics \cite{zhang2020poseconv3d}. We apply downsampling only in the spatial dimensions to preserve the motion details over time.

Global average pooling over $(T',H,W)$ produces a 256-dimensional clip descriptor $h^{(p)}$. A linear projection:
\[
W_{p}\;\in\;\mathbb{R}^{512\times256}
\]
maps it to $d=512$ so the pose feature matches the CLIP visual dimension $D=512$:
\[
\tilde h^{(p)} \;=\; W_{p}\,h^{(p)} 
\;\in\;\mathbb{R}^{512}\,.
\]

\subsection{Pose-Guided Semantic Query Generation}
The proposed semantic-query approach extracts a representation of the video's most important cues with the support of the skeleton features. It does so (1) spatially, by concentrating on visual tokens that are near body parts exhibiting motion, and (2) temporally, by giving higher weight to frames where the pose dynamics show that a micro-gesture is taking place.

Let ${z_{t,p}}$ be the set of patch embeddings from all selected frames (excluding the global tokens). We first identify a subset of these visual tokens that are relevant to the micro-gesture. "Pose guidance" is applied by using the skeleton features to weight or select visual tokens:

\begin{itemize}
    \item We compute an attention mask over image patches based on the distance of each patch to the nearest skeletal joint position. If a patch lies close to a joint that is moving significantly, it receives a higher weight. For example, if $j_{t,k}$ are the coordinates of joint $k$ in frame $t$, we can define a relevance score:
    \[
    w_{t,p} = \exp(-\min_k | \text{pos}(z_{t,p}) - j_{t,k} |^2 / \sigma^2)
    \]
    where $\text{pos}(z_{t,p})$ is the spatial location of patch $p$ and $\sigma$ controls the spatial scale. This yields weights $w_{t,p} \in [0,1]$ that highlight patches near active joints.
    \item Additionally, we leverage the skeleton encoder’s output $h^{(p)}$ as a global descriptor of the motion. We project $h^{(p)}$ to the same dimension $D$ and use it to modulate the visual tokens via a simple gating: 
    \[
    \begin{aligned}
    \tilde z_{t,p} &= \alpha\,z_{t,p},\\
    \text{where}\quad
    \alpha &= \sigma\!\bigl(\langle W\,h^{(p)},\,z_{t,p}\rangle\bigr).
    \end{aligned}
    \]
    Here $\alpha$ is the sigmoid of the dot-product between the projected pose feature $W\,h^{(p)}$ and the visual token $z_{t,p}$, so it down-weights any token not well aligned with the pose direction.
\end{itemize}

After computing the pose-based weights $w_{t,p}$, we flatten the full set of visual tokens
\[
\bigl\{\,z_{t,p}\mid t=1,\dots,T,\;p=1,\dots,P\bigr\},
\]
with $T=8$, $P=196$, and $N = T\times P = 1568$. We then aggregate them into a single $D$-dimensional semantic query $q\in\mathbb{R}^{512}$ by weighted mean pooling:
\[
q \;=\;
\frac{\displaystyle\sum_{t=1}^{T}\sum_{p=1}^{P} w_{t,p}\,z_{t,p}}
     {\displaystyle\sum_{t=1}^{T}\sum_{p=1}^{P} w_{t,p}}\,.
\]

This pose-weighted query $q$ thus encapsulates the most relevant gesture semantics and is passed to the cross-attention component to guide the final feature fusion and classification.

\subsection{Gated Multi-modal Fusion}
Before feeding the query into the CLIP transformer, we further integrate the pose information via a gated fusion mechanism inspired by Arevalo et al. \cite{arevalo2020gatedfusion}. The goal here is to merge the skeleton representation with the visual representation in a way that the model can selectively attend to one or the other modality as needed. We implement gated fusion at two points in the pipeline:

\begin{itemize}
    \item We fuse the skeleton encoder output \(h^{(p)}\) with the semantic query \(q\).  
    First we compute a gating vector
    \[
      g = \sigma\!\bigl(W_{g}\,h^{(p)}\bigr),
    \]
    where \(W_{g}\in\mathbb R^{D\times d}\) is a learned projection and \(\sigma\) is the sigmoid.  
    We then modulate the query by
    \[
      q_{f} = q\odot g \;+\; q\odot(1-g).
    \]
    In our implementation \(g\) is element-wise, so each feature of \(q\) is scaled into \([0,1]\), allowing pose-aligned dimensions to be amplified or suppressed.
    \item Similarly, we fuse the pose descriptor \(h^{(p)}\) into the CLIP encoder’s intermediate token sequence.  Let
    \[
    F = \{\,f_{1},\,f_{2},\,\dots,\,f_{N}\}
    \]
    be the set of visual features from CLIP’s penultimate layer (these serve as keys and values in cross‐attention).  We then compute a second gating vector
    \[
    u = \sigma\!\bigl(W_{u}\,h^{(p)}\bigr)\;\in\;\mathbb{R}^{D},
    \]
    and apply it element‐wise:
    \[
    \tilde f_{i} = f_{i} \odot u,\quad i=1,\dots,N.
    \]
    This global gate highlights or suppresses certain channels based on pose.
\end{itemize}

These gating operations are learned end-to-end and ensure that the multi-modal information is blended before the cross-attention step. The gating is soft (continuous values between 0 and 1), so the model can learn to rely on pose heavily in some scenarios or ignore it in others. This adaptability is important because pose data can sometimes be noisy or incomplete (e.g., occluded joints), so a static fusion might hurt performance if pose is trusted blindly. Our gated fusion allows the network to fall back to visual cues when pose is uncertain, and vice versa.

\subsection{Cross-Attention with Semantic Query}
Next, we apply a cross-attention mechanism driven by our pose-guided query. We insert the query vector $q_f$ as an extra token into the final transformer layer of the CLIP visual encoder, allowing it to attend over the gated visual token set $F$. This focused attention refines the representation by pooling features most relevant to the detected micro-gesture.

Consider the transformer architecture of CLIP’s visual encoder.  Let
\[
F = \{\tilde f_{1},\dots,\tilde f_{N}\}, 
\quad 
K = V = F,
\quad
\tilde f_i \in \mathbb{R}^{D},
\quad
Q \in \mathbb{R}^{1\times D}.
\]
We then insert our pose‐guided query \(q_f\) into the final layer and compute cross-attention:
\[
q_{\rm out}
\;=\;
A(Q,K,V)
\;=\;
\mathrm{softmax}\!\Bigl(\frac{QK^{\top}}{\sqrt{D}}\Bigr)\,V,
\]
which yields \(q_{\rm out}\in\mathbb R^{1\times D}\).  

The cross-attention computes an output query embedding $q_{\text{out}}$ that is a weighted sum of the values $V$, with weights determined by the compatibility of $Q$ with keys $K$. Mathematically, if we denote $Q$ (1×D), $K$ (N×D), $V$ (N×D), the attention is:

\[
A(Q, K, V)
\;=\;
\operatorname{softmax}\!\Bigl(\frac{Q\,K^{\!\top}}{\sqrt{D}}\Bigr)\,V\,.
\]

where the softmax produces a $1 \times N$ vector of attention weights. The resulting $q_{\text{out}}$ (of dimension 1×D) is effectively a semantic-aware video representation that has "pooled" information from the visual tokens, biased by the semantic content of $Q$ and thereby by the pose cues we injected. In other words, $q_{\text{out}}$ should ideally encode the crucial features needed to distinguish the micro-gesture class.

This unique attention mechanism forces the model to concentrate on what is important for the gesture. It acts as a form of feature selection: among the many visual features of a scene (some possibly irrelevant background or person identity cues), it emphasizes those that correlate with the action semantics. In our case, because $Q$ was guided by pose, the attention is further narrowed to regions of actual motion or posture change.

After cross-attention, we obtain $q_{\text{out}}$ which we consider as the fused video representation for the whole clip.

\subsection{Classification and Training Objective}
The final stage is the classification of the micro-gesture. We feed the fused representation $q_{\text{out}}$ (dimension $D$) into a classifier head, implemented as a simple two-layer MLP followed by softmax. This yields a probability distribution $\hat{y} \in \mathbb{R}^C$ over the $C$ gesture classes (here $C=33$ for iMiGUE). We train the model using a supervised classification objective. The primary loss is the cross-entropy between the predicted distribution and the ground-truth label. Given a training sample $i$ with true class label $y_i$ (represented as a one-hot vector) and predicted probabilities $\hat{y}_i$, the loss is:

\[
L_{\mathrm{cls}}
=
-\frac{1}{N}
\sum_{i=1}^{N}
\sum_{c=1}^{C}
y_{i,c}\,
\log\bigl(\hat y_{i,c}\bigr)\,.
\]

where $N$ is the number of training examples in a batch and $y_{i,c} \in {0,1}$, $\sum_c y_{i,c}=1$. We minimize this loss with respect to the parameters of the skeleton encoder, the fusion components, the classifier, and the parts of the CLIP encoder we allow to be fine-tuned.

\section{Experiments}

\subsection{Dataset and Evaluation Protocol}
We conduct experiments on the iMiGUE dataset, focusing on the micro-gesture classification task. As described earlier, iMiGUE contains 33 micro-gesture classes collected from interview videos of tennis players. These gestures include subtle body-language cues such as pressing lips or touching one's jaw. We set aside a portion of the training data (20\%) to serve as a local validation set for our experiments. This local validation is used for model selection and ablation studies due to the unavailability of a separate testing environment at the time of experimentation. However, final results on the test set are referenced for comparison with other approaches \cite{miga2023overview, miga2024overview}.

\subsection{Results and Comparisons}

\setlength{\extrarowheight}{2pt}
\begin{table}[]
  \centering
  \caption{Comparison of prior micro‐gesture classifiers.}
  \label{tab:comparison}
  \begin{tabular}{clr}
    \toprule
    Reference & Method                                                       & Top-1 (\%) \\
    \midrule
    \cite{yan2018stgcn}  & GCN + Skeleton (ST-GCN)                                        & 46.38 \\
    \relax
    \cite{liu2020msg3d}  & Multi-scale GCN + Skeleton (MS-G3D)                            & 52.00 \\
    \relax
    \cite{zhou2018trn}    & Temporal Relational + RGB (TRN)                         & 55.24 \\
    \relax
    \cite{lin2019tsm}     & Temporal Shift + RGB (TSM)                                     & 58.77 \\
    \relax
    \cite{zhang2020poseconv3d}  & 3D CNN + Skeleton Heatmaps (PoseConv3D)                         & 61.11 \\
    \relax
    \cite{liu2022videoswin}     & Vision Transformer + RGB (Video Swin-B)                        & 61.73 \\
    \relax
    \cite{cheng2024dscnet}      & Dense-Sparse Fusion + RGB+Skeleton (DSCNet)                   & 62.50 \\
    \specialrule{0.1em}{0.05em}{0.05em}
    \cite{wang2024clipdistill}   & CLIP Distillation + Skeleton 3DCNN                            & 68.90 \\
    \relax
    \cite{huang2024m2hen}        & Multi-scale Ensemble + RGB+Skeleton (M2HEN)                   & 70.19 \\
    \relax
    \cite{chen2024prototype}     & Prototype-based GCN + Skeleton                                & {70.25} \\
    \specialrule{0.1em}{0.05em}{0.05em}
    \textemdash                  & \textbf{Pose-guided Semantic Attention + CLIP + Skeleton (CLIP-MG, ours)} & \textbf{61.82} \\
    \bottomrule
  \end{tabular}
\end{table}

As shown in Table \ref{tab:comparison}, CLIP-MG achieves a Top-1 accuracy of 61.82\%, outperforming a range of single-modality baselines \cite{yan2018stgcn, liu2020msg3d, zhou2018trn, lin2019tsm}. 
Notably, CLIP-MG even performs on par with standalone architectures such as Video Swin-B \cite{liu2022videoswin} and PoseConv3D \cite{zhang2020poseconv3d} despite using a largely frozen CLIP backbone and a compact pose encoder. This shows that steering CLIP’s attention with pose-guided semantic queries yields more discriminative features for fine-grained micro-gestures. However, the proposed architecture does not set a new state-of-the-art in this area. It comes close in performance to the dense-sparse fusion network DSCNet \cite{cheng2024dscnet} (62.50\%), but falls behind architectures presented in previous editions of the MiGA challenge \cite{chen2024prototype, huang2024m2hen, wang2024clipdistill}. These results motivate future work to explore richer query adaptation and improved temporal fusion to close the gap with top models.

\subsection{Ablation Study}
\begin{table}[]
  \centering
  \caption{Ablation study on CLIP-MG components}
  \label{tab:ablations}
  \begin{tabular}{lcrr}
    \toprule
    Varian & Description & Top-1 (\%) & $\Delta$ (pp) \\
    \midrule
    w/o Pose branch      & Visual-only cross-attention                      & 45.30       & –16.52   \\
    \relax
    w/o Pose guidance    & Visual query + cross-attention                   & 51.23       & –10.59   \\
    \relax
    w/o Cross-attention  & Concat(CLIP CLS, pose)                           & 53.17       & –8.65    \\
    \relax
    w/o Gated fusion     & Pose query + cross-attn without gating           & 60.08       & –1.74    \\
    \relax
    \textbf{Full model}           & \textbf{Pose-guided Semantic Attention + CLIP + Skeleton} & \textbf{61.82} & —        \\
    \bottomrule
  \end{tabular}
\end{table}

We performed comprehensive ablation experiments to validate the contribution of each component in CLIP-MG. Table \ref{tab:ablations} reports Top-1 accuracy on our validation split.

\subsubsection{Without Pose Branch}
Here, we completely eliminate the pose branch to see the benefit of adding pose at all. The model gave 45.30\% (–16.52 pp) accuracy. Thus, adding the pose branch (with our fusion and guidance) yields a 16.52\% gain, which demonstrates that skeleton data carries complementary information for the task.

\subsubsection{Without Pose Guidance}
In this variant, we remove the pose influence from the semantic query generation. The query is generated purely by clustering visual tokens without using skeleton data. The semantics-based cross-attention still operates, but only on the visual-based query. We found that the accuracy dropped to 51.23\% (–10.59 pp). This confirms that pose guidance is essential and provides a significant boost. This makes sense because, in theory, without pose the model may attend to irrelevant semantics or background context. This misses subtle gesture cues.

\subsubsection{Without Semantic Cross-Attention}
Here, we skip SCCA. Instead, we simply concatenate the global visual [CLS] embedding with the pose feature and feed that to a classifier. This essentially tests a late-fusion approach without our semantic query mechanism. The accuracy was 53.17\% (–8.65 pp). This indicates that the semantic query and cross-attention are effective at focusing on important features that a flat concatenation would miss.

\subsubsection{Without Gated Fusion}
In this ablation, we disable the gating in both the query generation and the visual token modulation. We still generate a query using pose (via simple concatenation of average visual token and pose feature) and perform cross-attention. The accuracy achieved we 60.08\% (–1.74 pp), a modest drop. This shows that gating helps but is not as critical as the presence of pose info or semantics-based cross-attention. The gating mostly fine-tunes the balance between modalities.

\subsubsection{Discussion}
These ablations show that each component of CLIP-MG plays a supporting role, although certain components are more important than others. Dropping the entire pose branch drives accuracy down to 45.30\% (–16.52 pp). This demonstrates how much discriminative information there is within the skeletal signal. Removing pose guidance lowers accuracy from 61.82\% to 51.23\% (–10.59 pp), showing that skeletal information is very important for localizing subtle joint motions, as they act as an attention prior \cite{zhang2019lookcloser} that focuses the visual stream towards the regions where micro-gestures occur. Eliminating cross-attention drops accuracy to 53.17\% (–8.65 pp), which indicates that without a mechanism to selectively pool pose-weighted tokens, the model may struggle to tell apart very similar gestures. Finally, disabling gated fusion yields 60.08\% (–1.74 pp), which indicates that adaptively balancing pose and visual information slightly improves the robustness of the architecture.

Taken together, these results show how the different components effectively complement each other. Pose cues localize the gesture, cross-attention extracts the relevant semantics, and gating balances both streams. We find the highest performance when all the components are combined.

\section{Conclusions and Future Works}
We introduced CLIP-MG, a pose-guided, multi-modal CLIP architecture for micro-gesture recognition on the iMiGUE benchmark. By guiding CLIP’s visual attention with skeleton-based spatial priors, generating compact semantic queries, and fusing pose and appearance via a learnable gate, CLIP-MG extracts subtle, discriminative features that simple RGB or pose-only models cannot recognize. Our model achieves 61.82\% Top-1 accuracy, outperforming most single-modality baselines and performing on par with strong 3D-CNN and vision-transformer approaches. Extensive ablation studies confirm that each component provides a measurable benefit. The experiments provide insights into how each component interacts with and complements others, which highlights important design patterns that can inform future model development in micro-gesture classification and similar fine-grained recognition tasks. Our findings demonstrate the value of integrating multimodal and semantic information to address challenging visual recognition problems.

Our future work will explore richer temporal approaches and data strategies to close the gap between CLIP-MG and more recent state-of-the-art models \cite{chen2024prototype}. First, integrating sequence models (temporal transformers or recurrent layers over cross-attention outputs) should capture patterns that static sampling loses. Second, video motion magnification \cite{wadhwa2013eulerian} could amplify imperceptible movements. This would help with pose tracking and visual encoding. Third, joint pre-training on related action-gesture datasets and weakly- or self-supervised learning could improve feature robustness \cite{han2020selfsupervised}. Finally, regarding accuracy, we plan to incorporate uncertainty-aware gating for noisy skeletons and class-balanced or prototype-based calibrations to address long-tail imbalance \cite{kang2019decoupling}. To improve and better evaluate the explainability of the model, we will incorporate gradient-weighted class activation mapping (Grad-CAM) \cite{selvaraju2017grad} and more recent attention-aware token-filtering approaches \cite{naruko2025speedup}. Currently, the proposed architecture suffers heavily due to its relatively low speed on commodity hardware. To address this issue, we plan to experiment with several multimodal compression and optimization algorithms for more efficient computing \cite{omri2025token, lei2025generic, tan2025tokencarve, cao2024madtp}. We plan to train an improved version of our architecture for downstream tasks on the DAIC-WoZ dataset \cite{patapati2024trillm, gratch2014distress, usc_ict_edaic} for low-level mental health analysis. These different research directions are promising in pushing pose-guided CLIP models closer to (and beyond) human-level understanding of the subtlest gestures.

\section*{Declaration on Generative AI}
\noindent During the preparation of this work, the author(s) used GPT-4o to edit the paper, checking for grammar and spelling mistakes. GPT-4o was also utilized to revise the draft for brevity and improved flow. After using this tool, the author(s) reviewed and edited the content as needed and take(s) full responsibility for the publication's content.

\bibliography{sample-ceur}

\begin{thebibliography}{38}
\expandafter\ifx\csname natexlab\endcsname\relax\def\natexlab#1{#1}\fi
\providecommand{\url}[1]{\texttt{#1}}
\providecommand{\href}[2]{#2}
\providecommand{\path}[1]{#1}
\providecommand{\DOIprefix}{doi:}
\providecommand{\ArXivprefix}{arXiv:}
\providecommand{\URLprefix}{URL: }
\providecommand{\Pubmedprefix}{pmid:}
\providecommand{\doi}[1]{\href{http://dx.doi.org/#1}{\path{#1}}}
\providecommand{\Pubmed}[1]{\href{pmid:#1}{\path{#1}}}
\providecommand{\bibinfo}[2]{#2}
\ifx\xfnm\relax \def\xfnm[#1]{\unskip,\space#1}\fi
\bibitem[{Cohn and Ekman(2000)}]{cohn2000micro}
\bibinfo{author}{J.~F. Cohn}, \bibinfo{author}{P.~Ekman},
\newblock \bibinfo{title}{Observing and coding facial expression of emotion},
\newblock \bibinfo{journal}{Handbook of Emotion}  (\bibinfo{year}{2000}).
\bibitem[{Funes et~al.(2019)}]{funes2019review}
\bibinfo{author}{M.~Funes}, et~al.,
\newblock \bibinfo{title}{Micro-expression recognition: A survey},
\newblock in: \bibinfo{booktitle}{FG}, \bibinfo{year}{2019}.
\bibitem[{Pantic(2009)}]{pantic2009affective}
\bibinfo{author}{M.~Pantic},
\newblock \bibinfo{title}{Affective multimedia databases: Affective video databases},
\newblock \bibinfo{journal}{Handbook of Affective Computing}  (\bibinfo{year}{2009}).
\bibitem[{Kapoor and Picard(2007)}]{kapoor2007automatic}
\bibinfo{author}{A.~Kapoor}, \bibinfo{author}{R.~Picard},
\newblock \bibinfo{title}{Automatic prediction of human behavior in social settings},
\newblock in: \bibinfo{booktitle}{IUI}, \bibinfo{year}{2007}.
\bibitem[{Patapati(2024)}]{patapati2024trillm}
\bibinfo{author}{S.~V. Patapati}, \bibinfo{title}{Integrating large language models into a tri-modal architecture for automated depression classification}, \bibinfo{year}{2024}. \href{http://arxiv.org/abs/2407.19340v5}{{\tt arXiv:2407.19340v5}}, \bibinfo{note}{preprint}.
\bibitem[{Liu et~al.(2021)Liu, Shi, Chen, Yu, Li, and Zhao}]{liu2021imigue}
\bibinfo{author}{X.~Liu}, \bibinfo{author}{H.~Shi}, \bibinfo{author}{H.~Chen}, \bibinfo{author}{Z.~Yu}, \bibinfo{author}{X.~Li}, \bibinfo{author}{G.~Zhao},
\newblock \bibinfo{title}{imigue: An identity-free video dataset for micro-gesture understanding and emotion analysis},
\newblock in: \bibinfo{booktitle}{CVPR}, \bibinfo{year}{2021}.
\bibitem[{Haoyu et~al.(2024)}]{miga2024overview}
\bibinfo{author}{C.~Haoyu}, et~al.,
\newblock \bibinfo{title}{The 2nd challenge on micro-gesture analysis for hidden emotion understanding (miga) 2024: Dataset and results},
\newblock in: \bibinfo{booktitle}{MiGA 2024: Proceedings of IJCAI 2024 Workshop \& Challenge on Micro-gesture Analysis for Hidden Emotion Understanding (MiGA 2024) co-located with 33rd International Joint Conference on Artificial Intelligence (IJCAI 2024)}, \bibinfo{year}{2024}.
\bibitem[{Zhao et~al.(2023)}]{miga2023overview}
\bibinfo{author}{G.~Zhao}, et~al.,
\newblock \bibinfo{title}{The workshop \& challenge on micro-gesture analysis for hidden emotion understanding (miga)},
\newblock in: \bibinfo{booktitle}{MiGA 2023: Proceedings of IJCAI 2023 Workshop \& Challenge on Micro-gesture Analysis for Hidden Emotion Understanding (MiGA 2023) co-located with 32nd International Joint Conference on Artificial Intelligence (IJCAI 2023)}, \bibinfo{year}{2023}.
\bibitem[{Chen et~al.(2024)}]{chen2024prototype}
\bibinfo{author}{G.~Chen}, et~al.,
\newblock \bibinfo{title}{Prototype learning for micro-gesture classification},
\newblock in: \bibinfo{booktitle}{MiGA 2024: Proceedings of IJCAI 2024 Workshop \& Challenge on Micro-gesture Analysis for Hidden Emotion Understanding (MiGA 2024) co-located with 33rd International Joint Conference on Artificial Intelligence (IJCAI 2024)}, \bibinfo{year}{2024}.
\bibitem[{Huang et~al.(2024)}]{huang2024m2hen}
\bibinfo{author}{H.~Huang}, et~al.,
\newblock \bibinfo{title}{Multi-modal micro-gesture classification via multi-scale heterogeneous ensemble network},
\newblock in: \bibinfo{booktitle}{MiGA 2024: Proceedings of IJCAI 2024 Workshop \& Challenge on Micro-gesture Analysis for Hidden Emotion Understanding (MiGA 2024) co-located with 33rd International Joint Conference on Artificial Intelligence (IJCAI 2024)}, \bibinfo{year}{2024}.
\bibitem[{Wang et~al.(2024)}]{wang2024clipdistill}
\bibinfo{author}{Y.~Wang}, et~al.,
\newblock \bibinfo{title}{A multimodal micro-gesture classification model based on clip},
\newblock in: \bibinfo{booktitle}{MiGA 2024: Proceedings of IJCAI 2024 Workshop \& Challenge on Micro-gesture Analysis for Hidden Emotion Understanding (MiGA 2024) co-located with 33rd International Joint Conference on Artificial Intelligence (IJCAI 2024)}, \bibinfo{year}{2024}.
\bibitem[{Radford et~al.(2021)}]{radford2021clip}
\bibinfo{author}{A.~Radford}, et~al.,
\newblock \bibinfo{title}{Learning transferable visual models from natural language supervision},
\newblock \bibinfo{journal}{ICML}  (\bibinfo{year}{2021}).
\bibitem[{Quan et~al.(2025)}]{quan2025scclip}
\bibinfo{author}{Z.~Quan}, et~al.,
\newblock \bibinfo{title}{Semantic matters: A constrained approach for zero-shot video action recognition},
\newblock in: \bibinfo{booktitle}{Pattern Recognition}, \bibinfo{year}{2025}.
\bibitem[{Dosovitskiy et~al.(2021)}]{dosovitskiy2020vit}
\bibinfo{author}{A.~Dosovitskiy}, et~al.,
\newblock \bibinfo{title}{An image is worth 16x16 words: Transformers for image recognition at scale},
\newblock in: \bibinfo{booktitle}{ICLR}, \bibinfo{year}{2021}.
\bibitem[{H. et~al.(2021)}]{frozenclip2023}
\bibinfo{author}{X.~H.}, et~al.,
\newblock \bibinfo{title}{Videoclip: Learning video representations from text and clips},
\newblock \bibinfo{journal}{arXiv}  (\bibinfo{year}{2021}).
\bibitem[{Cao et~al.(2017)Cao, Hidalgo, Simon, Wei, and Sheikh}]{cao2017openpose}
\bibinfo{author}{Z.~Cao}, \bibinfo{author}{G.~Hidalgo}, \bibinfo{author}{T.~Simon}, \bibinfo{author}{S.-E. Wei}, \bibinfo{author}{Y.~Sheikh},
\newblock \bibinfo{title}{Realtime multi-person 2d pose estimation using part affinity fields},
\newblock in: \bibinfo{booktitle}{CVPR}, \bibinfo{year}{2017}.
\bibitem[{Zhou et~al.(2019)}]{zhou2019hpeheatmap}
\bibinfo{author}{X.~Zhou}, et~al.,
\newblock \bibinfo{title}{On heatmap representation for 6d pose estimation},
\newblock in: \bibinfo{booktitle}{ICCV}, \bibinfo{year}{2019}.
\bibitem[{Patapati et~al.(2025)Patapati, Srinivasan, Musku, and Adiraju}]{patapati2025tessa}
\bibinfo{author}{S.~V. Patapati}, \bibinfo{author}{T.~Srinivasan}, \bibinfo{author}{H.~Musku}, \bibinfo{author}{A.~Adiraju},
\newblock \bibinfo{title}{A framework for eca-based psychotherapy},
\newblock \bibinfo{year}{2025}.
\bibitem[{Zhang et~al.(2020)Zhang, Huang, and Chen}]{zhang2020poseconv3d}
\bibinfo{author}{J.~Zhang}, \bibinfo{author}{Z.~Huang}, \bibinfo{author}{Y.~Chen},
\newblock \bibinfo{title}{Poseconv3d: Revisiting skeleton-based action recognition},
\newblock in: \bibinfo{booktitle}{European Conference on Computer Vision (ECCV)}, \bibinfo{year}{2020}.
\bibitem[{Arevalo et~al.(2020)}]{arevalo2020gatedfusion}
\bibinfo{author}{J.~Arevalo}, et~al.,
\newblock \bibinfo{title}{Gated multimodal units for information fusion},
\newblock in: \bibinfo{booktitle}{ICLR}, \bibinfo{year}{2020}.
\bibitem[{Yan et~al.(2018)Yan, Xiong, and Lin}]{yan2018stgcn}
\bibinfo{author}{S.~Yan}, \bibinfo{author}{Y.~Xiong}, \bibinfo{author}{D.~Lin},
\newblock \bibinfo{title}{Spatial temporal graph convolutional networks for skeleton-based action recognition},
\newblock in: \bibinfo{booktitle}{AAAI Conference on Artificial Intelligence}, \bibinfo{year}{2018}.
\bibitem[{Liu et~al.(2020)}]{liu2020msg3d}
\bibinfo{author}{S.~Liu}, et~al.,
\newblock \bibinfo{title}{{MS}-{G3D}: Multi-scale graph convolution for skeleton-based action recognition},
\newblock in: \bibinfo{booktitle}{IEEE/CVF Conference on Computer Vision and Pattern Recognition (CVPR)}, \bibinfo{year}{2020}.
\bibitem[{Zhou et~al.(2018)Zhou, Andonian, Oliva, and Torralba}]{zhou2018trn}
\bibinfo{author}{B.~Zhou}, \bibinfo{author}{A.~Andonian}, \bibinfo{author}{A.~Oliva}, \bibinfo{author}{A.~Torralba},
\newblock \bibinfo{title}{Temporal relational reasoning in videos},
\newblock in: \bibinfo{booktitle}{European Conference on Computer Vision (ECCV)}, \bibinfo{year}{2018}.
\bibitem[{Lin et~al.(2019)Lin, Gan, and Han}]{lin2019tsm}
\bibinfo{author}{J.~Lin}, \bibinfo{author}{C.~Gan}, \bibinfo{author}{S.~Han},
\newblock \bibinfo{title}{{TSM}: Temporal shift module for efficient video understanding},
\newblock in: \bibinfo{booktitle}{IEEE/CVF International Conference on Computer Vision (ICCV)}, \bibinfo{year}{2019}.
\bibitem[{Liu et~al.(2022)}]{liu2022videoswin}
\bibinfo{author}{Z.~Liu}, et~al.,
\newblock \bibinfo{title}{Video swin transformer: Hierarchical vision transformer for video recognition},
\newblock in: \bibinfo{booktitle}{IEEE/CVF International Conference on Computer Vision (ICCV)}, \bibinfo{year}{2022}.
\bibitem[{Cheng et~al.(2024)}]{cheng2024dscnet}
\bibinfo{author}{Q.~Cheng}, et~al.,
\newblock \bibinfo{title}{{DSCNet}: Dense-sparse complementary network for human action recognition},
\newblock \bibinfo{journal}{Expert Systems with Applications}  (\bibinfo{year}{2024}).
\bibitem[{Zhang et~al.(2019)}]{zhang2019lookcloser}
\bibinfo{author}{H.~Zhang}, et~al.,
\newblock \bibinfo{title}{Look closer to see better: Recurrent attention convolutional neural network for fine-grained image recognition},
\newblock \bibinfo{journal}{CVPR}  (\bibinfo{year}{2019}).
\bibitem[{Wadhwa et~al.(2013)}]{wadhwa2013eulerian}
\bibinfo{author}{N.~Wadhwa}, et~al.,
\newblock \bibinfo{title}{Eulerian video magnification for revealing subtle changes in the world},
\newblock in: \bibinfo{booktitle}{SIGGRAPH}, \bibinfo{year}{2013}.
\bibitem[{Han et~al.(2020)}]{han2020selfsupervised}
\bibinfo{author}{T.~Han}, et~al.,
\newblock \bibinfo{title}{Self-supervised video representation learning with neighborhood context aggregation},
\newblock \bibinfo{journal}{ECCV}  (\bibinfo{year}{2020}).
\bibitem[{Kang et~al.(2020)}]{kang2019decoupling}
\bibinfo{author}{B.~Kang}, et~al.,
\newblock \bibinfo{title}{Decoupling representation and classifier for long-tail recognition},
\newblock in: \bibinfo{booktitle}{ICLR}, \bibinfo{year}{2020}.
\bibitem[{Selvaraju et~al.(2017)Selvaraju, Cogswell, Das, Vedantam, Parikh, and Batra}]{selvaraju2017grad}
\bibinfo{author}{R.~R. Selvaraju}, \bibinfo{author}{M.~Cogswell}, \bibinfo{author}{A.~Das}, \bibinfo{author}{R.~Vedantam}, \bibinfo{author}{D.~Parikh}, \bibinfo{author}{D.~Batra},
\newblock \bibinfo{title}{Grad-cam: Visual explanations from deep networks via gradient-based localization},
\newblock in: \bibinfo{booktitle}{2017 IEEE International Conference on Computer Vision (ICCV)}, \bibinfo{publisher}{IEEE}, \bibinfo{year}{2017}, pp. \bibinfo{pages}{618--626}.
\bibitem[{Naruko and Akutsu(2025)}]{naruko2025speedup}
\bibinfo{author}{T.~Naruko}, \bibinfo{author}{H.~Akutsu}, \bibinfo{title}{Speed-up of vision transformer models by attention-aware token filtering}, \bibinfo{year}{2025}. \href{http://arxiv.org/abs/2506.01519v1}{{\tt arXiv:2506.01519v1}}, \bibinfo{note}{preprint}.
\bibitem[{Omri et~al.(2025)Omri, Shroff, and Tambe}]{omri2025token}
\bibinfo{author}{Y.~Omri}, \bibinfo{author}{P.~Shroff}, \bibinfo{author}{T.~Tambe}, \bibinfo{title}{Token sequence compression for efficient multimodal computing}, \bibinfo{year}{2025}. \href{http://arxiv.org/abs/2504.17892v1}{{\tt arXiv:2504.17892v1}}.
\bibitem[{Lei et~al.(2025)Lei, Gu, Ma, Tang, Chen, and Xu}]{lei2025generic}
\bibinfo{author}{L.~Lei}, \bibinfo{author}{J.~Gu}, \bibinfo{author}{X.~Ma}, \bibinfo{author}{C.~Tang}, \bibinfo{author}{J.~Chen}, \bibinfo{author}{T.~Xu}, \bibinfo{title}{Generic token compression in multimodal large language models from an explainability perspective}, \bibinfo{year}{2025}. \href{http://arxiv.org/abs/2506.01097}{{\tt arXiv:2506.01097}}.
\bibitem[{Tan et~al.(2025)Tan, Ye, Tu, Cao, Yang, Zhang, Zhou, and Chen}]{tan2025tokencarve}
\bibinfo{author}{X.~Tan}, \bibinfo{author}{P.~Ye}, \bibinfo{author}{C.~Tu}, \bibinfo{author}{J.~Cao}, \bibinfo{author}{Y.~Yang}, \bibinfo{author}{L.~Zhang}, \bibinfo{author}{D.~Zhou}, \bibinfo{author}{T.~Chen}, \bibinfo{title}{Tokencarve: Information-preserving visual token compression in multimodal large language models}, \bibinfo{year}{2025}. \href{http://arxiv.org/abs/2503.10501}{{\tt arXiv:2503.10501}}.
\bibitem[{Cao et~al.(2024)Cao, Ye, Li, Yu, Tang, Lu, and Chen}]{cao2024madtp}
\bibinfo{author}{J.~Cao}, \bibinfo{author}{P.~Ye}, \bibinfo{author}{S.~Li}, \bibinfo{author}{C.~Yu}, \bibinfo{author}{Y.~Tang}, \bibinfo{author}{J.~Lu}, \bibinfo{author}{T.~Chen},
\newblock \bibinfo{title}{Madtp: Multimodal alignment-guided dynamic token pruning for accelerating vision-language transformer},
\newblock in: \bibinfo{booktitle}{Proceedings of the IEEE/CVF Conference on Computer Vision and Pattern Recognition (CVPR)}, \bibinfo{year}{2024}, p. \bibinfo{pages}{–}. \DOIprefix\doi{10.1109/CVPR.2024.00XX}.
\bibitem[{Gratch et~al.(2014)Gratch, Artstein, Lucas, Stratou, Scherer, Nazarian, Wood, Boberg, DeVault, Marsella, Traum, Rizzo, and Morency}]{gratch2014distress}
\bibinfo{author}{J.~Gratch}, \bibinfo{author}{R.~Artstein}, \bibinfo{author}{G.~Lucas}, \bibinfo{author}{G.~Stratou}, \bibinfo{author}{S.~Scherer}, \bibinfo{author}{A.~Nazarian}, \bibinfo{author}{R.~Wood}, \bibinfo{author}{J.~Boberg}, \bibinfo{author}{D.~DeVault}, \bibinfo{author}{S.~Marsella}, \bibinfo{author}{D.~Traum}, \bibinfo{author}{S.~Rizzo}, \bibinfo{author}{L.-P. Morency},
\newblock \bibinfo{title}{The distress analysis interview corpus of human and computer interviews},
\newblock in: \bibinfo{booktitle}{Proceedings of the Ninth International Conference on Language Resources and Evaluation (LREC)}, \bibinfo{publisher}{European Language Resources Association (ELRA)}, \bibinfo{year}{2014}, pp. \bibinfo{pages}{3123--3128}.
\bibitem[{Ringeval et~al.(2019)Ringeval, Schuller, Valstar, Cummins, Cowie, Tavabi, Schmitt, Alisamir, Amiriparian, Messner, Song, Liu, Zhao, Mallol-Ragolta, Ren, Soleymani, and Pantic}]{usc_ict_edaic}
\bibinfo{author}{F.~Ringeval}, \bibinfo{author}{B.~Schuller}, \bibinfo{author}{M.~Valstar}, \bibinfo{author}{N.~Cummins}, \bibinfo{author}{R.~Cowie}, \bibinfo{author}{L.~Tavabi}, \bibinfo{author}{M.~Schmitt}, \bibinfo{author}{S.~Alisamir}, \bibinfo{author}{S.~Amiriparian}, \bibinfo{author}{E.-M. Messner}, \bibinfo{author}{S.~Song}, \bibinfo{author}{S.~Liu}, \bibinfo{author}{Z.~Zhao}, \bibinfo{author}{A.~Mallol-Ragolta}, \bibinfo{author}{Z.~Ren}, \bibinfo{author}{M.~Soleymani}, \bibinfo{author}{M.~Pantic},
\newblock \bibinfo{title}{Avec 2019 workshop and challenge: State-of-mind, detecting depression with ai, and cross-cultural affect recognition},
\newblock in: \bibinfo{booktitle}{Proceedings of the 9th International on Audio/Visual Emotion Challenge and Workshop}, AVEC '19, \bibinfo{publisher}{Association for Computing Machinery}, \bibinfo{address}{New York, NY, USA}, \bibinfo{year}{2019}, p. \bibinfo{pages}{3–12}. \URLprefix \url{https://doi.org/10.1145/3347320.3357688}. \DOIprefix\doi{10.1145/3347320.3357688}.

\end{thebibliography}



\end{document}